\documentclass[conference,a4paper]{IEEEtran}
\IEEEoverridecommandlockouts
\usepackage{microtype}

\usepackage[hidelinks]{hyperref}
\usepackage[cmex10]{amsmath}%American Math Society(AMS) math formatting
\usepackage{amssymb,amsfonts}%AMS extra symbols and fonts
\usepackage{dblfloatfix}%fix double column figure ordering and placement

\usepackage[ruled,vlined]{algorithm2e}
\usepackage{graphicx}
\graphicspath{{Figures/PDF/}{Figures/PNG/}}

\usepackage{booktabs}
\usepackage{siunitx}
\usepackage[numbers,compress]{natbib}
\usepackage{texnames}
\usepackage{bm,bbm}
\usepackage{orcidlink}
\usepackage{xcolor}
\definecolor{cvprblue}{rgb}{0.21,0.49,0.74}
\usepackage{hyperref}

\usepackage{mathtools}

\usepackage{multirow}

\hypersetup{
  colorlinks=false,
  pdfborder={0 0 1},
  linkbordercolor={0 0 1}
}

\usepackage[absolute,overlay]{textpos}
\newcommand{\ieeecopyrightnotice}{%
  \begin{textblock*}{\textwidth}(18mm,8mm)
    \noindent{\footnotesize\raggedright\copyright~2026 IEEE. Personal use of this material is permitted. Permission from IEEE must be obtained for all other uses, in any current or future media, including reprinting/republishing this material for advertising or promotional purposes, creating new collective works, for resale or redistribution to servers or lists, or reuse of any copyrighted component of this work in other works.\par}
  \end{textblock*}}
  
\begin{document}

\title{\uppercase{Set-Based Transformer for Atmospheric Compensation in Standoff LWIR Hyperspectral Imaging}
%\thanks{funding from AFOSR?}
}

\author{
\IEEEauthorblockN{
Fabian Perez\orcidlink{0009-0005-9655-1708},
Nicolas Quintero\orcidlink{0009-0006-0079-632X},
Jeferson Acevedo\orcidlink{0009-0008-0840-1972},
Hoover Rueda-Chacón\orcidlink{0000-0002-6763-8629}
}
\IEEEauthorblockA{
\textit{Department of Computer Science, Universidad Industrial de Santander}\\
Bucaramanga, 680002, Colombia\\
\{perez2258059,nicolas2220090,jeferson2221790\}@correo.uis.edu.co, hfarueda@uis.edu.co
}
}

\maketitle
\ieeecopyrightnotice

\begin{abstract}
	Passive long-wave infrared (LWIR) hyperspectral imaging under a standoff geometry depends on atmospheric absorption and emission, as well as reflected radiance, thus making atmospheric compensation essential to get knowledge of a target of interest. Despite its importance, this compensation has been largely overlooked due to its practical and modeling difficulty. In this paper, we present a lightweight set-based deep learning framework that takes multiple radiance measurements, collected at different standoff ranges, as input and jointly estimates transmittance, atmospheric path radiance, and a shared downwelling spectrum. We analyze the learned representation with a sparse autoencoder and observe that several latent features do activate on geographically coherent subsets of the test data despite the absence of location supervision. Experiments on a MODTRAN generated standoff LWIR dataset demonstrate low spectral distortion across all estimated products. 
    The dataset and code is publicly available at: \url{https://factral.co/SAE-LWIR/}

\end{abstract}

 
\section{Introduction}

Unlike the visible to shortwave infrared (VIS–SWIR) range, which is dominated by reflected solar radiation (0.4–2.5 µm) \cite{articlevis-swir}, the long-wave infrared (LWIR) senses thermally emitted radiation within the 8–14 µm atmospheric window \cite{manolakis2019longwave}. This enables sensing capabilities independent of solar illumination, thus making LWIR suitable for day–night operation. As a result, LWIR has been widely used in security and surveillance \cite{Rodger2016LWIR,Muller2013LWIR}, autonomous driving and infrastructure monitoring \cite{LI2021357,Judd2019LWIR}, wildfire detection \cite{Johnston2014LWIR}, environmental remote sensing \cite{FENG2018340}, and hyperspectral target detection \cite{Hackwell1996LWIR}.

\begin{figure}[!t]
    \centering
    \includegraphics[width=1\linewidth]{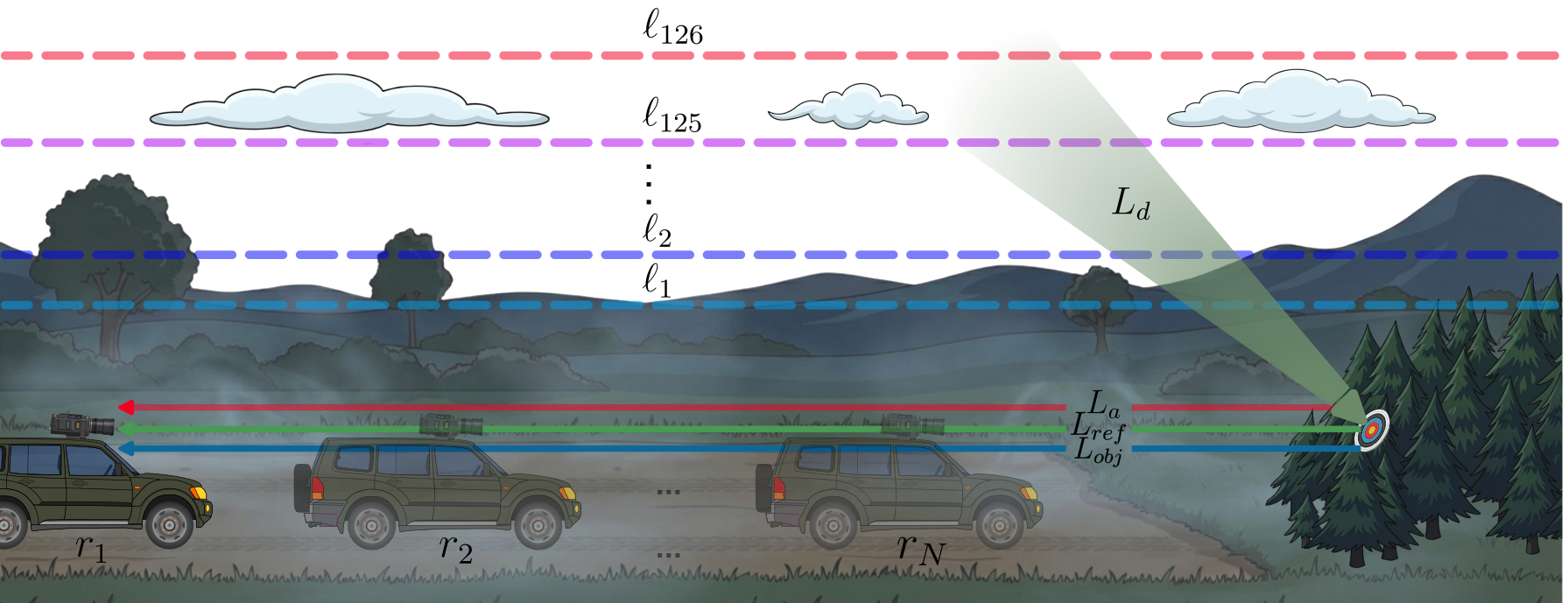}
    \caption{\textbf{Standoff LWIR imaging configuration}. The atmosphere is discretized into 126 layers. Downwelling irradiance from the sky $L_d$ illuminates the target. The at-sensor radiance at the hyperspectral LWIR camera is shown as the sum of (i) target thermal emission $L_{obj}$, (ii) reflected downwelling irradiance $L_{ref}$, and (iii) atmospheric path emission along the line of sight $L_a$} 
    %The camera size is exaggerated for clarity.}
    \label{fig:configuration}
\end{figure}

LWIR measurements acquired under a standoff sensing configuration are governed by the radiative transfer equation (RTE) and are affected by atmospheric attenuation through the path transmittance $\tau$, atmospheric absorption and self-emission $L_a$ (upwelling), as well as by surface reflection $L_{ref}$ driven by downwelling $L_d$. The forward standoff model, following the RTE in~\cite{manolakis2019longwave, Thome1998atmospheric}, is given by
\begin{equation}
L(\lambda; r_n; T)
=
\tau(\lambda;r_n)
\Big[
\underbrace{\varepsilon\,B(\lambda; T)}_{L_{\mathrm{obj}}}
+
\underbrace{\rho \, L_d(\lambda)}_{L_{\mathrm{ref}}}
\Big]
+ L_{a}(\lambda;r_n).
\label{eq:forward}
\end{equation}
where $L(\lambda; r_n; T)$ denotes the at-sensor radiance at wavelength $\lambda$ for the $n$-th standoff range (distance) $r_n$ and target temperature $T$, $\varepsilon$ is the object emmisivity and $\rho$ is the surface albedo. Unlike airborne and satellite remote sensing, standoff acquisition operates at shorter ranges where the same scene may be observed across multiple standoff distances, making atmospheric effects range-dependent as depicted in \autoref{fig:configuration}. %The measured at-sensor radiance combines three terms: 
%The resulting LWIR measurement combines three terms: attenuated target emission, attenuated reflected downwelling atmospheric radiance, and atmospheric path radiance as depicted in \autoref{fig:configuration}; 
Hence, accurate downstream tasks on the target require atmospheric compensation (AC), which demands the estimation of transmittance, upwelling, and downwelling (TUD).

\begin{figure*}[!t]
    \centering
    \centerline{\includegraphics[width=\linewidth]{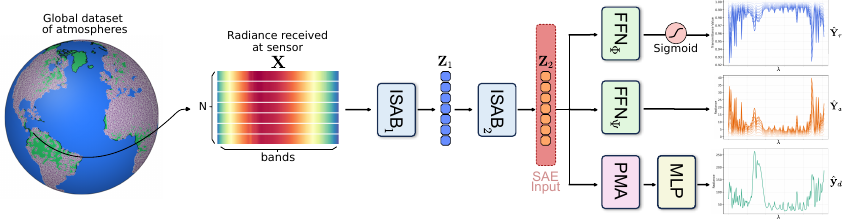}}
\caption{\textbf{Proposed Framework}. The input consists of $N$ radiance measurements selected from the globally generated dataset by sampling a single location (pink dots on the globe) and extracting its observations at $N$ different standoff ranges; each sample has $B=256$ spectral bands beetween $8\mu$m and $13\mu$m, stacked as $\mathbf{X}\in\mathbb{R}^{N\times B}$. A Set-Transformer encoder with two ISAB modules encode the set of measurements into a latent representation. The latent features are then routed to three task-specific decoders: an FFN with a sigmoid activation predicts atmospheric transmittance $\hat{\mathbf{Y}}_\tau\in\mathbb{R}^{N\times B}$; an FFN predicts atmospheric path radiance $\hat{\mathbf{Y}}_a\in\mathbb{R}^{N\times B}$; a PMA module followed by an MLP aggregates across the set to estimate a shared downwelling spectrum $\hat{\mathbf{y}}_d\in\mathbb{R}^{B}$.}
    
    \label{fig:method}
\end{figure*}

%Despite its importance, atmospheric compensation has historically been overlooked in standoff sensing scenarios. Stelter et al. \cite{stelter2024atmospheric} model long-range atmospheric absorption but rely on externally estimated parameters and ignore range-diverse measurements. Bao et al. \cite{bao2023heat} propose a physics-based correction approach, yet assume known environmental conditions and do not address ill-posed near-horizontal paths. Gallastegi et al. \cite{gallastegi2025absorption} analyze absorption-induced distortions, but require prior atmospheric characterization and ignore joint estimation across shared atmospheric paths. Classical LWIR atmospheric correction methods for airborne or spaceborne imaging rely on calibration targets or auxiliary atmospheric data, limiting their applicability to ground-based standoff scenarios.

Building on this, the AC literature has largely focused on airbone and satellite rather than standoff sensing. In near-nadir viewing geometry, the upwelling radiance and transmittance can be treated as approximately the same for all pixels~\cite{pieper2016scene}, which simplifies estimation. State-of-the-art works like \cite{chen2018retrieval} proposed a neural network to estimate atmospheric components jointly with land-surface temperature, but the method is tailored to satellite nadir observations. In \cite{westing2020multimodal}, a variational autoencoder (VAE) learning framework is proposed, again restricted to the nadir regime. Oblique remote configurations, where upwelling and transmittance vary across the field of view have been explored in \cite{o2022oblique}, but assume known range or rely on MODTRAN~\cite{10.1117/12.578758} (MODerate resolution atmospheric TRANsmission) look-up tables for AC estimation. Similarly, \cite{xu2020multiple} proposes a convolutional neural network (CNN) to estimate atmospheric products (TUD estimation) but evaluates only remote configurations.

In standoff LWIR, near-horizontal and range-diverse paths make the radiative terms strongly geometry-dependent, and the problem remains comparatively underexplored. In \cite{stelter2024atmospheric}, a diffusion-based framework models long-range absorption for VIS/NIR/SWIR but does not address LWIR. In \cite{bao2023heat}, a CNN-based approach is proposed but neglects the atmospheric path-radiance contribution. In \cite{gallastegi2025absorption}, absorption is leveraged to estimate depth, while omitting the reflected downwelling term in the RTE. Collectively, these limitations highlight the need for AC methods that explicitly handle LWIR standoff imaging.
%, including range diversity and the full radiative-transfer components required for accurate downstream inference.

In this work, we propose a lightweight set-based deep learning framework for atmospheric compensation in passive standoff LWIR imaging. Here, set-based refers to jointly processing multiple radiance measurements acquired at different standoff ranges, without assuming any ordering \cite{NIPS2017_f22e4747}. By jointly exploiting multiple radiance measurements acquired at different standoff ranges, the method estimates the following AC products: atmospheric transmittance ($\tau$), upwelling radiance ($L_a$), and a shared downwelling spectrum ($L_d$). The main contributions are listed as follows:
\begin{itemize}
\item A set-based framework for joint TUD estimation in passive standoff LWIR imaging, coupled with a sparse autoencoder to enable latent-space interpretability.
\item The first publicly available MODTRAN-generated dataset for atmospheric compensation in standoff LWIR imaging.
\end{itemize}

\section{Proposed Method}
We propose a set-based framework to estimate AC products from standoff hyperspectral LWIR observations as illustrated in \autoref{fig:method}. Each sample input consist of $N$ measurements acquired at different standoff ranges, each represented as a spectrum over $B$ spectral bands. We first describe the neural network that predicts the AC products, and then introduce a sparse autoencoder that exposes structure in the learned latent representations.

\begin{figure}[!b]
    \centering
    \includegraphics[width=1\linewidth]{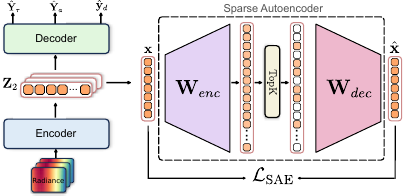}
\caption{Sparse Autoencoder pipeline. Encoder tokens $\mathbf{Z}_2$ are the input to the SAE that reconstructs token activations by minimizing $\mathcal{L}_{\mathrm{SAE}}$.}
    \label{fig:placeholder}
\end{figure}

\subsection{Set-based Neural Network}

Let the $n$-th standoff measurement, acquired at range $r_n$, be a radiance spectrum $\mathbf{x}_n \in \mathbb{R}^{B}$, and let the input set be written as
$\mathbf{X} \in \mathbb{R}^{N \times B}$ whose rows are the elements $\{\mathbf{x}_n\}_{n=1}^{N}$ collected at distinct standoff ranges $\mathcal{R}=\{r_n\}_{n=1}^{N}$. Note that multiple ranges provide additional constraints that mitigate the ill-posed problem of estimating the (AC) products, since $\tau$ and $L_a$ vary with $r_n$. 
The network predicts three outputs:
a range-wise transmittance $\hat{\mathbf{Y}}_\tau \in \mathbb{R}^{N \times B}$,
a range-wise upwelling $\hat{\mathbf{Y}}_a \in \mathbb{R}^{N \times B}$,
and a shared downwelling spectrum $\hat{\mathbf{y}}_d \in \mathbb{R}^{B}$.
We denote the three task-specific branches as
\begin{equation}
\hat{\mathbf{Y}}_\tau = \mathcal{G}_T(\mathbf{X}), \qquad
\hat{\mathbf{Y}}_a = \mathcal{G}_U(\mathbf{X}), \qquad
\hat{\mathbf{y}}_d = \mathcal{F}_d(\mathbf{X}).
\end{equation}

\textbf{Symmetry requirements.}
Because the $N$ measurements form a set, their ordering is arbitrary.
Let $\mathbf{P}\in\{0,1\}^{N\times N}$ be a permutation matrix and denote the row-permuted input by $\mathbf{P}\cdot X$. Accordingly,
%Let $\pi$ denote a permutation of $N$ elements, and let $\pi \cdot \mathbf{X}$ indicate permuting the rows of $\mathbf{X}$ accordingly.
%The transmittance and upwelling branches must be \emph{permutation equivariant}, since they preserve a one-to-one mapping between each input measurement and its corresponding output:
$\hat{\mathbf{Y}}_\tau$ and $\hat{\mathbf{Y}}_a$ are predicted per standoff range, so each output must stay aligned with the same input;
%As a result, if we permute the input ordering, the outputs should be permuted in exactly the same way; 
thus, each branch must be permutation equivariant, that is,
\begin{equation}
\mathcal{G}_T(\mathbf{P} \cdot \mathbf{X}) = \mathbf{P} \cdot \mathcal{G}_T(\mathbf{X}), 
\qquad
\mathcal{G}_U(\mathbf{P} \cdot \mathbf{X}) = \mathbf{P} \cdot \mathcal{G}_U(\mathbf{X}).
\end{equation}
In contrast, downwelling is shared across the $N$ standoff ranges, for a fixed atmosphere, and therefore must be \emph{permutation invariant}, which can be denoted as
\begin{equation}
\mathcal{F}_d(\mathbf{P} \cdot \mathbf{X}) = \mathcal{F}_d(\mathbf{X}).
\end{equation}
\textbf{Set encoder.} To satisfy the requirements described above, we adopt a Set-Transformer encoder~\cite{lee2019set}.
%and map the input set to latent tokens
%$\mathbf{Z} \in \mathbb{R}^{N \times d}$, where each row $\mathbf{z}_n \in \mathbb{R}^{d}$ is a latent token associated with the $n$-th measurement.
Specifically, we use ISAB (\textit{Induced Set Attention Block}) which is built from a Multihead Attention Block (MAB)~\cite{vaswani2017attention}.
%
%which computes attention using queries $\mathbf{Q}$, keys $\mathbf{K}$, and values $\mathbf{V}$. Let $d_k$ denote the dimensionality of the key vectors per attention head. The attention operation is defined as:
%\begin{equation}
%\mathrm{Att}(\mathbf{Q},\mathbf{K},\mathbf{V}) \;=\;
%\mathrm{softmax}\!\left(\frac{\mathbf{Q}\mathbf{K}^\top}{\sqrt{d_k}}\right)\mathbf{V}.
%\end{equation}
%Multihead attention with $h$ heads is given by
%\begin{equation}
%\begin{aligned}
%\mathrm{Multihead}(\mathbf{Q},\mathbf{K},\mathbf{V})
%&=
%\mathrm{Concat}\big(\mathrm{head}_1,\ldots,\mathrm{head}_h\big)\mathbf{W}^O,\\
%\mathrm{head}_i
%&=
%\mathrm{Att}(\mathbf{Q}\mathbf{W}_i^Q,\mathbf{K}\mathbf{W}_i^K,\mathbf{V}\mathbf{W}_i^V),
%\end{aligned}
%\end{equation}
%and MAB applies multihead attention followed by a token-wise feed-forward sublayer with residual connections and normalization:
%\begin{equation}
%\begin{aligned}
%\mathbf{H}&=\mathrm{LN}\!\Big(\mathbf{X}+\mathrm{Multihead}(\mathbf{X},\mathbf{Y},\mathbf{Y})\Big),\\
%\mathrm{MAB}(\mathbf{X},\mathbf{Y})
%&=
%\mathrm{LN}\!\Big(\mathbf{H} + \mathrm{FF}\big(\mathbf{H}\big)\Big).
%\end{aligned}
%\end{equation}
Using $m$ inducing points $\mathbf{I} \in \mathbb{R}^{m \times d}$, ISAB introduces a compact set of learnable latent \textit{anchors} that attend to the input set to form a global summary $\mathbf{S}$, and then lets each element in $\mathbf{X}$ attend to this summary~\cite{lee2019set}. 
%This mechanism preserves permutation equivariance while reducing the attention cost by routing interactions through $m \ll N$ inducing points~\cite{lee2019set}.
ISAB can be expressed as:
\begin{align}
\mathrm{ISAB}_m(\mathbf{X}) &= \mathrm{MAB}(\mathbf{X}, \mathbf{S}) \in \mathbb{R}^{N \times d}, \label{eq:isab1}\\
\text{where}\quad \mathbf{S} &= \mathrm{MAB}(\mathbf{I}, \mathbf{X}) \in \mathbb{R}^{m \times d}. \label{eq:isab2}
\end{align}
We use two stacked ISAB layers to increase the depth of the set encoder so as to capture higher-order interactions
\begin{equation}
\mathbf{Z}_2 = \mathrm{ISAB}_2\big(\mathrm{ISAB}_1(\mathbf{X})\big).
\end{equation}
ISAB is \textit{permutation equivariant} by construction, so permuting the input permutes the latent tokens in the same way~\cite{lee2019set}.

%Beyond symmetry, ISAB replaces full self-attention over $N$ elements (quadratic in $N$) with attention mediated by $m$ inducing points, reducing complexity to $O(Nm)$ while still allowing each measurement to interact with the rest of the set through global context. This is especially attractive in standoff LWIR, where the measurements at different ranges are coupled through shared atmospheric structure~\cite{manolakis2019longwave} and the model benefits from explicitly modeling cross-range dependencies in latent space.

\textbf{Equivariant transmittance and upwelling decoders.}
The transmittance and upwelling branches apply the same feed-forward network (FFN) independently to each latent token. Concisely, we use two token-wise FFNs, $\mathrm{FFN}_{\phi}$ and $\mathrm{FFN}_{\psi}$, each mapping $\mathbb{R}^{d}\!\to\!\mathbb{R}^{B}$ and applied to the encoder output $\mathbf{Z}_2$:
\begin{equation}
\hat{\mathbf{Y}}_\tau = \sigma\!\big(\mathrm{FFN}_{\phi}(\mathbf{Z}_2)\big), \qquad
\hat{\mathbf{Y}}_a = \mathrm{FFN}_{\psi}(\mathbf{Z}_2),
\end{equation}
where $\sigma(\cdot)$ is an element-wise sigmoid that enforces $\hat{\mathbf{Y}}_\tau \in (0,1)$, and $\mathrm{FFN}_{\phi},\mathrm{FFN}_{\psi}$ to share the same structure but to not share weights across tasks.
For the $n$-th token $\mathbf{z}_n$, we define the token-wise FFN output by
\begin{equation}
\big[\mathrm{FFN}(\mathbf{Z}_2)\big]_n \;:=\; \mathrm{FFN}(\mathbf{z}_n),
\qquad
\mathbf{z}_n = [\mathbf{Z}_2]_n.
\end{equation}
The FFN computes
\begin{equation}
\begin{aligned}
\mathbf{h}^{(1)}_n &= \mathrm{LN}_1\!\Big(\mathrm{GELU}(\mathbf{W}_1\mathbf{z}_n+\mathbf{b}_1)\Big),\\
\mathbf{h}^{(2)}_n &= \mathrm{LN}_2\!\Big(\mathrm{GELU}(\mathbf{W}_2\mathbf{h}^{(1)}_n+\mathbf{b}_2)\Big),
\end{aligned}
\end{equation}
where $\mathrm{LN}$ denotes layer normalization~\cite{ba2016layer} and $\mathrm{GELU}$ refer to the activation function~\cite{hendrycks2016gaussian}, and outputs
\begin{equation}
\mathbf{o}_n = \mathbf{W}_3\mathbf{h}^{(2)}_n+\mathbf{b}_3, \quad\mathbf{o}_n   \in \mathbb{R}^{B}.
\end{equation}
Stacking $\{\mathbf{o}_n\}_{n=1}^{N}$ yields $\mathrm{FFN}(\mathbf{Z}_2) \in \mathbb{R}^{N\times B}$.
Applying the FFN to every token makes the output permutation-equivariant.
%\begin{equation}
%\mathrm{FFN}(\pi \cdot \mathbf{Z}_2) = \pi \cdot \mathrm{FFN}(\mathbf{Z}_2).
%\end{equation}
%This guarantees the required one-to-one correspondence for both $\hat{\mathbf{T}}$ and $\hat{\mathbf{U}}$.

\textbf{Invariant downwelling decoder.}
We assume downwelling is common to all the measurements in a set, since they are acquired under the same atmosphere; therefore, we aggregate $\mathbf{Z}_2$ into a single vector using Pooling by Multihead Attention (PMA)~\cite{lee2019set}. PMA introduces $k$ learned seed vectors $\mathbf{S}\in\mathbb{R}^{k\times d}$ and pools the set through attention as implemented by
\begin{equation}
\mathbf{v} = \mathrm{PMA}(\mathbf{Z}_2) :=\mathrm{MAB}(\mathbf{S}, \mathbf{Z}_2) \in \mathbb{R}^{k\times d}.
\end{equation}
We use $k{=}1$ so that the pooled embedding is $\mathbf{v} \in \mathbb{R}^{d}$. We then post-process this pooled representation with a linear projection to obtain the $B$-band downwelling estimate:
\begin{align}
    &\hat{\mathbf{y}}_d \;=\; \mathrm{MLP}\big(\mathbf{v}\big), \quad \hat{\mathbf{d}}\in \mathbb{R}^{B}, \\
    &\text{where}  \quad \mathrm{MLP}(\mathbf{v}) \;:=\; \mathbf{W}_d\,\mathbf{v} + \mathbf{b}_d,
\end{align}
%
%
%
%Because PMA performs attention-based pooling over the set, the resulting mapping is permutation invariant.
%
%
%
%

\textbf{Training objective.}
We train the network by minimizing the mean squared error (MSE) between the predicted atmospheric products $(\hat{\mathbf{Y}}_\tau,\hat{\mathbf{Y}}_a,\hat{\mathbf{y}}_d)$ and their corresponding ground-truth targets $(\mathbf{Y}_\tau,\mathbf{Y}_a,\mathbf{y}_d)$.
%Let  denote the  %transmittance, upwelling, and downwelling 
%associated with the input set $\mathbf{X}$, and let  be the network outputs. 
We define the per-sample losses as
\begin{equation}
\begin{aligned}
\mathcal{L}_{\tau} &= \left\lVert \hat{\mathbf{Y}}_\tau - \mathbf{Y}_\tau \right\rVert_2^{2}, \qquad
\mathcal{L}_{a} = \left\lVert \hat{\mathbf{Y}}_a - \mathbf{Y}_a \right\rVert_2^{2},\\
&\hspace{1.4cm}\mathcal{L}_{d} = \left\lVert \hat{\mathbf{y}}_d - \mathbf{y}_d \right\rVert_2^{2},
\end{aligned}
\end{equation}
and optimize the total objective $\mathcal{L}$, 
\begin{equation}
\mathcal{L} = \mathcal{L}_{\tau} + \mathcal{L}_{a} + \mathcal{L}_{d}.
\end{equation}
\vspace{-4mm}
\subsection{Sparse Autoencoder (SAE)}
To better understand \emph{what} the trained model encodes and whether its internal features align with physically meaningful factors, we analyze the frozen trained set encoder using a sparse autoencoder (SAE)~\cite{makhzani2013k}. 
%SAEs have recently once again emerged as an effective tool for interpreting neural representations by learning a sparse dictionary of latent features that can isolate coherent directions in activation space~\cite{king2025leveraging}.
Concretely, given an input set $\mathbf{X}$, the encoder outputs $\mathbf{Z}_2 \in \mathbb{R}^{N \times d}$, and we treat each token $\mathbf{z}_n = [\mathbf{Z}_2]_n \in \mathbb{R}^{d}$ as an activation vector to be reconstructed with a sparse, overcomplete representation as depicted in \autoref{fig:placeholder}.

Let $\mathbf{x}\in\mathbb{R}^{d}$ denote a token activation, i.e., $\mathbf{x}=\mathbf{z}_n$. The SAE encoder produces sparse features $\mathbf{a}\in\mathbb{R}^{M}$ using TopK gating~\cite{makhzani2013k}:
\begin{equation}
\mathbf{a} 
:= \mathrm{TopK}\!\big(\mathbf{W}_{\mathrm{enc}}\mathbf{x}+\mathbf{b}_{\mathrm{enc}}\big),
\qquad
\mathbf{W}_{\mathrm{enc}}\in\mathbb{R}^{M\times d},
\end{equation}
 where $\mathrm{TopK}(\cdot)$ retains only the $K$ largest-magnitude entries and sets the remaining activations to zero. The decoder then reconstructs the token activation with a linear dictionary:
\begin{equation}
\hat{\mathbf{x}} = \mathbf{a}^{\top}\mathbf{W}_{\mathrm{dec}} + \mathbf{b}_{\mathrm{dec}},
\qquad
\mathbf{W}_{\mathrm{dec}}\in\mathbb{R}^{M\times d},\;
\mathbf{b}_{\mathrm{dec}}\in\mathbb{R}^{d}.
\end{equation}
We train the SAE using the per-token $\ell_2$ reconstruction loss:
\begin{equation}
\mathcal{L}_{\mathrm{SAE}}(\mathbf{x}) = \left\lVert \mathbf{x} - \hat{\mathbf{x}} \right\rVert_2^{2}.
\end{equation}
%After training, the sparse codes $\mathbf{a}$ provide an interpretable basis for analyzing the latent patterns.

\section{Dataset Generation}

%\FPcomment{
%\begin{itemize}
%    \item what we need, tramittance, upwelling downelling for the standoff configuration, and then the forward we assume
%    \item about modtran and the version 5, its capabilities, and what we develop to use it (pymodtran etc)
%    \item details about our configuration for the generation of downelling (hemispherical, 45 degrees), upwelling (setting the target temperature to 0), etc, info about CSP
%    \item size of the dataset, variations, fixed parameters in the forward, train,val,test sizes, etc.
%\end{itemize}
%}

\begin{table}[!b]
\centering
\caption{Test-set performance for different set sizes $N$.}
\label{tab:test_results}
\begin{tabular}{c l c c}
\hline
\textbf{$N$} & \textbf{Target} & \textbf{SAM} $\downarrow$ & \textbf{NRMSE} $\downarrow$ \\
\hline
\multirow{3}{*}{1}
& Transmittance & 0.0057 & 0.0554 \\
& Upwelling     & 0.1244 & 0.0564 \\
& Downwelling   & 0.1937 & 0.1740 \\
\hline
\multirow{3}{*}{7}
& Transmittance & \textbf{0.0025} & \textbf{0.0093} \\
& Upwelling     & \textbf{0.0330} & \textbf{0.0093} \\
& Downwelling   & \textbf{0.0409} & \textbf{0.0193} \\
\hline
\end{tabular}
\end{table}

To the best of our knowledge, there is currently no publicly available dataset tailored to AC in standoff LWIR imaging. To address this gap, we generate a large-scale simulated dataset using the clear-sky atmospheric profile database (CSP)~\cite{rs14102329}. This database is derived from the \textit{European Centre for Medium-Range Weather Forecast Reanalysis V5 }(ERA5) and contains 82,828 profiles with key atmospheric state variables, including latitude, longitude, pressure, temperature, specific humidity and ozone concentration at 136 layers $l_n$. We further filter these profiles by discarding cases with cloud coverage exceeding 10\% or relative humidity above 90\%, and by removing a subset of ocean-surface profiles, thus yielding 36,547 clear-sky profiles.
Radiance at-sensor was simulated using seven target temperatures $T$ ranging from $280 K$ to $310 K$ inclusive, assuming a fixed gray-body like emissivity $\varepsilon$ value of $0.95$.

% The forward standoff model, following the radiative transfer equation in~\cite{manolakis2019longwave}, is given by
% %
% \begin{equation}
% L(\lambda; r_n; T_0)
% =
% \tau(\lambda;r_n)
% \Big[
% \underbrace{\varepsilon\,B(\lambda; T_0)}_{L_{\mathrm{obj}}}
% +
% \underbrace{\rho \, L_d(\lambda)}_{L_{\mathrm{ref}}}
% \Big]
% + L_{a}(\lambda;r_n).
% \end{equation}

% %
% where $L(\lambda; r_n; T_0)$ denotes the at-sensor radiance at wavelength $\lambda$ for the $n$-th standoff range $r_n$ and target temperature $T_0$. 
% %For compactness, we use 
% %$\tau_n(\lambda)\coloneqq \tau(\lambda; r_n)$ and $u_n(\lambda)\coloneqq u(\lambda; r_n)$.

MODTRAN5~\cite{10.1117/12.578758}, accessed through an internally developed Python tool, was used to calculate atmospheric products and the at-sensor radiance following~\autoref{eq:forward}.
%: transmittance ($\tau$), upwelling radiance ($L_a$), and downwelling radiance ($L_d$), for each profile. 
Specifically, it computes band-integrated at-sensor radiance convolving with the instrumental spectral response function (ISRF), and we report radiance in microflicks $(\mu\mathrm{W}\cdot \mathrm{sr}^{-1}\cdot \mathrm{cm}^{-2}\cdot \mu\mathrm{m}^{-1})$. Gas concentrations were specified independently for each of the first 126 layers $\ell_{126}$ (maximum number in MODTRAN5) using the filtered database. Downwelling radiance $L_d$ was calculated at a fixed viewing angle of $45^{\circ}$ from the target as an approximated representation of the hemispherical sky radiance incident on the target. We define $N=7$ and $\mathcal{R}=\{30,\,90,\,150,\,210,\,270,\,330,\,390\}\,\mathrm{m}$. The atmospheric path radiance $L_a$ (upwelling) and transmittance $\tau$ were computed along the line of sight for each $r_n$, as illustrated in Fig.~\ref{fig:configuration}.

The resulting standoff MODTRAN-generated dataset has 36,547 profiles, 7 ranges, and 7 temperatures, yielding a total of 255,829 samples. The dataset was randomly split into 70\% for training, 10\% for validation, and 20\% for testing.

\begin{figure}[!b]
    \centering
    \includegraphics[width=1\linewidth]{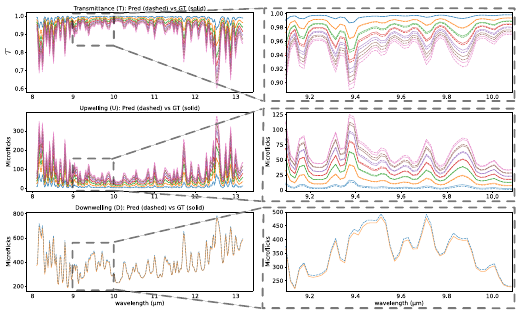}
    \caption{Qualitative results on one test sample predicted (dashed) and ground-truth (solid) are shown for transmittance $\mathbf{Y}_\tau$, atmospheric path radiance $\mathbf{Y}_a$, and downwelling radiance $\mathbf{y}_d$ across the LWIR window. The left column shows the full spectral range, while the right column provides a zoomed view.}
    \label{fig:tud}
\end{figure}
\section{Results and Analysis}

We evaluate on the test split of the generated standoff LWIR dataset. We use $N=1$ with $r_n=270$m and $N=7$ standoff measurements per sample, and an encoder embedding size of $d=256$. The spectral dimension is set to $B=256$, consistent with the bands of a real standoff sensor, the DARPA LWIR hyperspectral sensor~\cite{gallastegi2025absorption,yellin2024concurrent} over the $8$--$13\,\mu\mathrm{m}$ window; we model each channel spectral response as a Gaussian with $\mathrm{FWHM}=40\,\mathrm{nm}$. Training is performed for 510k iterations with batch size 512 and learning rate $1\times10^{-3}$, using AdamW with weight decay 0.01. All experiments are run on a single NVIDIA RTX PRO 6000 Blackwell GPU; training requires only $\sim$10\% of the available VRAM, indicating that the proposed architecture is lightweight and scalable.~\autoref{tab:test_results} reports test-set performance for the three estimated AC products. We report Spectral Angle Mapper (SAM) and normalized RMSE (NRMSE). As expected, performance improves substantially from $N=1$ to $N=7$, since single-measurement estimation is inherently ill-posed while range-diverse measurements better constrain the solution. Overall, the model achieves low spectral distortion across all AC products, with particularly strong agreement for transmittance.
\autoref{fig:tud} shows qualitative predictions for one randomly selected test sample, including the estimated $\hat{\mathbf{Y}}_\tau$, $\hat{\mathbf{Y}}_a$, and $\hat{\mathbf{y}}_d$ compared against their corresponding ground-truths.
The plotted spectra highlight that the network preserves fine-grained spectral structure.

\begin{figure}[!t]
    \centering
    \includegraphics[width=0.9\linewidth]{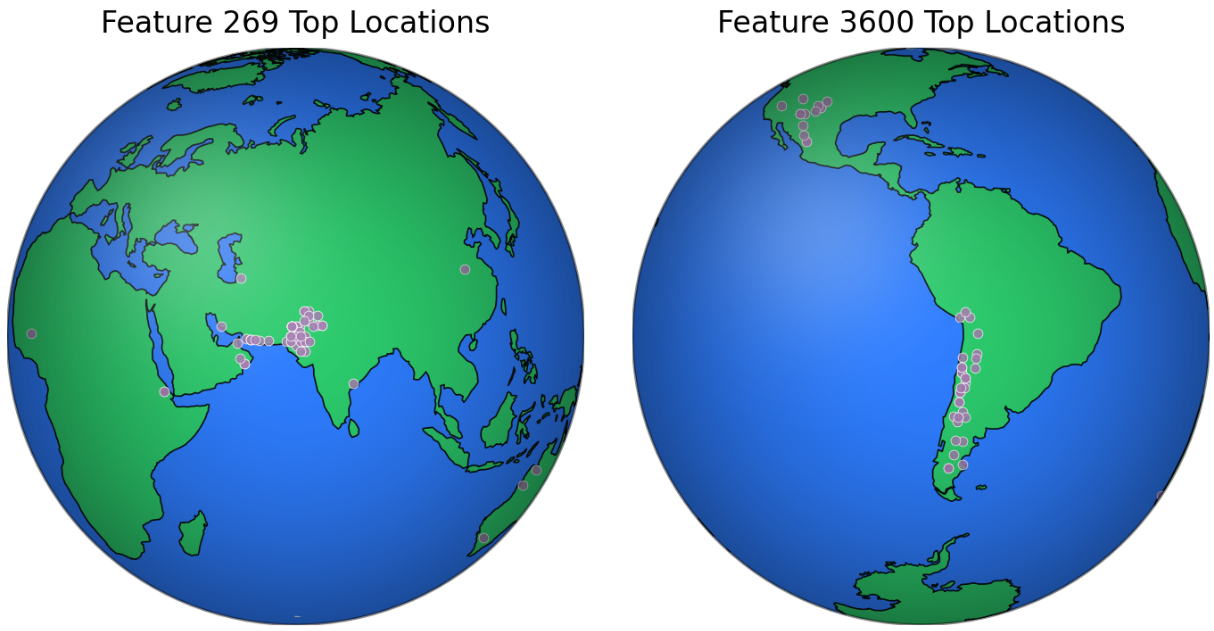}
    \caption{Top-activating locations for two SAE features. Each point denotes a test sample among the highest activations.}
    \label{fig:sae_top_examples}
\end{figure}

\textbf{SAE analysis.}
To better understand the latent space learned by the encoder, we train a SAE on the frozen activations $\mathbf{Z}_2 \in \mathbb{R}^{N \times d}$. We use a sparse feature dimension of $M=3072$ (i.e., $12\times d$) to decouple latent factors, and we optimize the SAE for 20k iterations using AdamW. The TopK gating parameter is set to $k=16$, selected via cross-validation. After training, we perform a top-activating analysis~\cite{gao2024scaling} to identify the samples that most strongly activate each sparse feature. Interestingly, for multiple features, the highest-activation examples consistently correspond to radiance sets originating from highly similar geographic regions, despite the fact that the network is never provided with explicit location. This emergent behavior suggests that the model organizes the features by clustering atmospheres with similar physical conditions,
%, leading to location-consistent latent factors.
\autoref{fig:sae_top_examples} plots the top-activating examples for two sparse features; note how the activations form clear location-related clusters.

%\FPcomment{
%\begin{itemize}
%\item MSE and SAM results for the test set in the proposed network
%\item chosen hyperparameters for the SAE and its performance
%\end{itemize}
%}

%FPcomment{
%\begin{itemize}
%\item some cool plots comparing GT and predictions
%\item analyze failure cases?
%\end{itemize}
%}

%\subsection{Sparse Feature Interpretability}
%\FPcomment{
%\begin{itemize}
%\item show top samples activation per selected features and its similarities between each top samples and give some interpretation 
%\item find and show a specific feature that represent something physically meaningful
%\item show a t-sne analysis?
%\end{itemize}
%}
\section{Conclusions}
We presented a set-based deep learning framework for standoff LWIR atmospheric compensation that estimates transmittance, atmospheric path radiance, and a shared downwelling radiance from range-diverse measurements. We showed that SAE provides a powerful interpretability tool for probing physically grounded representations. Future work will model angle-dependent downwelling, incorporate spectrally varying emissivity, and study sensitivity to the number of measurements and the selected standoff ranges, while exploring alternative SAE variants and model capacities.

\vspace{-0.7mm}
\section{Acknowledgments}
This work was supported by the Air Force Office of Scientific Research (AFOSR) through the Southern Office of Aerospace Research and Development (SOARD) under grant number FA8655-25-1-8010

\small
\bibliographystyle{IEEEtranN}
\bibliography{references}

\end{document}